\documentclass[letterpaper, 10 pt, conference]{ieeeconf}  

\IEEEoverridecommandlockouts                              
\usepackage{graphicx}
\usepackage{float}
\usepackage{caption}
\usepackage{subcaption}
\usepackage{multicol}
\usepackage{multirow}
\usepackage{hyperref}

\usepackage{amsmath,bm}
\usepackage{amssymb}
\usepackage{indentfirst}
\usepackage{graphicx}
\usepackage{wrapfig}
\usepackage{color,soul} 
\newcommand{\scalemath}[2]{\scalebox{#1}{\mbox{\ensuremath{\displaystyle #2}}}}
\definecolor{red}{rgb}{1.00,0.00,0.00}
\definecolor{blue}{rgb}{0.00,0.00,1.00}
\definecolor{green}{rgb}{0.30, 0.50,0.00}

\newcommand{\cblue}[1] {\textcolor{blue}{#1}}

\usepackage{flushend} 

\title{
\LARGE \bf Harnessing the Synergy between Pushing, Grasping, and Throwing \\to Enhance Object Manipulation in Cluttered Scenarios
}

\author{Hamidreza Kasaei$^{1*}$ and Mohammadreza Kasaei$^{2*}$
\thanks{$^*$ Equal contribution}%
\thanks{$^{1}$Hamidreza~Kasaei is with the Department of Artificial Intelligence, University of Groningen, The Netherlands. Email: hamidreza.kasaei@rug.nl}%
\thanks{$^{2}$ Mohammadreza Kasaei is with the School of Informatics, University of Edinburgh, UK. Email: m.kasaei@ed.ac.uk}%
\thanks{\cblue{This paper has been accepted at ICRA 2024.}}
}

\begin{document}

\maketitle
\thispagestyle{empty}
\pagestyle{empty}

\begin{abstract}

In this work, we delve into the intricate synergy among non-prehensile actions like pushing, and prehensile actions such as grasping and throwing, within the domain of robotic manipulation. We introduce an innovative approach to learning these synergies by leveraging model-free deep reinforcement learning. The robot's workflow involves detecting the pose of the target object and the basket at each time step, predicting the optimal push configuration to isolate the target object, determining the appropriate grasp configuration, and inferring the necessary parameters for an accurate throw into the basket. This empowers robots to skillfully reconfigure cluttered scenarios through pushing, creating space for collision-free grasping actions. Simultaneously, we integrate throwing behavior, showcasing how this action significantly extends the robot's operational reach. Ensuring safety, we developed a simulation environment in Gazebo for robot training, applying the learned policy directly to our real robot. Notably, this work represents a pioneering effort to learn the synergy between pushing, grasping, and throwing actions. 
Extensive experimentation in both simulated and real-robot scenarios substantiates the effectiveness of our approach across diverse settings. 
Our approach achieves a success rate exceeding 80\% in both simulated and real-world scenarios.
A video showcasing our experiments is available online at: \href{https://youtu.be/q1l4BJVDbRw}{https://youtu.be/q1l4BJVDbRw}
\end{abstract}
\section{Introduction}
\begin{figure*}[!t]
\centering
\includegraphics[width=\linewidth]{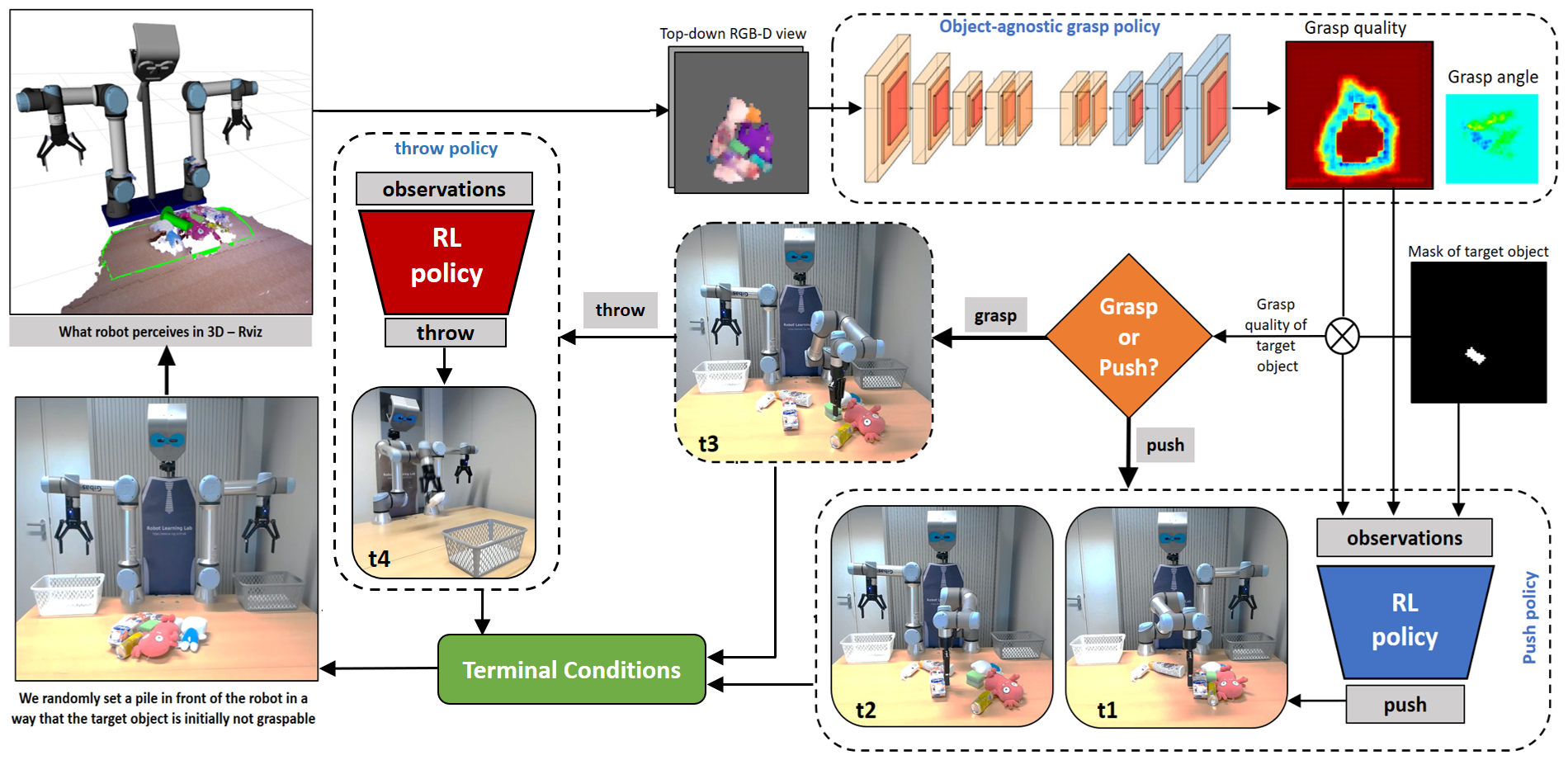}
\caption{Overview: The perception system provides essential inputs, including top-down RGB-D views of the workspace, and a mask highlighting the target object. These inputs are then processed through an object-agnostic grasp policy, resulting in pixel-wise grasp synthesis for the scene. Based on the grasp quality of the target object, the system makes a decision between executing a push action or a grasp action. Specifically, if the grasp quality surpasses a predefined threshold, the robot initiates a grasp; otherwise, it proceeds with a push action. After successfully grasping the target object, the robot leverages throwing actions when the target basket is out of its immediate reach.}
\vspace{-4mm}
\label{fig:overview}
\end{figure*}

The field of robotic manipulation has evolved significantly,
 focusing on the synergy between non-prehensile actions like pushing and prehensile actions such as grasping and throwing~\cite{zhou2023learning,zeng2018learning}.  The importance of learning such a synergy became clear in cluttered scenarios. Pushing, grasping, and throwing are not just individual actions; they are interconnected skills that, when combined, unlock a world of possibilities for service robots. Our approach integrates these actions into a unified framework to enable robots to autonomously manage clutter and execute complex tasks. 
 This involves a sequence of training steps, starting with an object-agnostic grasping foundation, followed by strategic push, and then throw policies. 
 We adopt a modular training approach,
 allowing easier troubleshooting and updates without impacting the whole system. Moreover, this approach reduces the need for extensive data collection for end-to-end training, as each policy is optimized separately but coherently. Particularly, the push policy is conditioned on the output of an object-agnostic grasping policy, ensuring a structured framework essential for complex manipulation tasks. To mitigate the safety risks and time-intensive nature of real-world training, we developed a simulation environment in Gazebo that closely mirrors our real robot.
 An overview of the proposed method is depicted in Fig.~\ref{fig:overview}.
Through simulation training and empirical evaluation, our framework proves to be both adaptable and effective, achieving superior performance compared to baseline alternatives.
Our key contributions are groundbreaking in three main areas. Firstly, we are pioneers in integrating pushing, grasping, and throwing actions to expand the capabilities of robots in complex environments. Secondly, our approach, although trained on simulation data, shows remarkable adaptability and generalization in real-world scenarios. Lastly, our framework achieves a success rate exceeding 80\% in both simulated and real-world scenarios, confirming its practical viability. 

\section{Related Work}

In this section, we provide a brief overview of the related work, organizing it into several distinct subsections. 

\noindent\textbf{Pushing techniques}: Numerous studies have contributed to the advancement of pushing strategies for efficiently reorganizing cluttered objects. Previous research has encompassed analytical techniques \cite{yu2016more,moura2022non} as well as learning-based approaches \cite{bauza2018data,lowrey2018reinforcement, dengler2022learning, florence2022implicit,ferrandis2023nonprehensile} for push planning. These investigations have significantly improved object rearrangement in intricate settings. In contrast to these approaches, our method conditions the push policy on the grasp quality map of the scene. This implies that the robot learns to execute a push action that renders the target object graspable.

\noindent\textbf{Grasping methods:} Grasping, as a cornerstone of robotic manipulation, has seen remarkable progress, with a spectrum of techniques ranging from analytical grasp synthesis to data-driven learning methods. Notably, object-agnostic grasping approaches have garnered attention for their adaptability across diverse object types and scenarios \cite{kumra2020antipodal,morrison2018closing,kumra2020antipodal,kasaei2023mvgrasp}. We encourage readers to explore a comprehensive survey on object grasping \cite{newbury2023deep} for a more in-depth understanding of this field. Among all possible object-grasping approaches, we adopted the MVGRASP~\cite{kasaei2023mvgrasp} approach as our grasp policy network in this work.
 
\noindent\textbf{Synergy between pushing and grasping:}
The domains of pushing and grasping have been the focal points of extensive research within the field of robotic manipulation \cite{zhou2023learning,zhang2023reinforcement,zuo2023graph,zeng2018learning}. The synergy between these two fundamental actions has led to significant advancements in the realm of object manipulation. 
These studies have explored how pushing can pave the way for successful grasps, thereby enhancing efficiency and mitigating collisions in cluttered environments \cite{zhou2023learning,zhang2023reinforcement,zuo2023graph,zeng2018learning,zuo2023graph}. In contrast to these approaches, our method prioritizes learning to grasp objects in cluttered scenarios before acquiring a push policy to render the target object graspable. Moreover, our approach stands out in its ability to simultaneously learn the length and direction of the push in a single pass. This differs from most state-of-the-art approaches, which typically require passing a batch of 16 images to the network and subsequently identifying the image with the highest q-value~\cite{zhou2023learning,zeng2018learning}. The direction of the push is determined based on the index of the image, while the push's length remains a fixed value. 


\noindent\textbf{Throwing object:} Early works in the field of robotic throwing primarily focused on precise object placement. These studies often involved analytical approaches and dedicated throwing kernels~\cite{hu2010ball,gai2013motion,lofaro2012humanoid}. The challenge lay in accurately predicting throwing parameters to achieve desired placements, a task further complicated by varying object shapes and environmental constraints \cite{zeng2020tossingbot,kasaei2023throwing}. Recent advancements in deep reinforcement learning have offered new avenues for teaching robots the art of throwing~\cite{zeng2020tossingbot,kasaei2023throwing}. Similar to these approaches, we used an RL-based method to enable robots to adapt and generalize their throwing capabilities.
Integrating throwing with pushing and grasping actions represents a novel and promising direction. 

\section{Method} 
\label{problemformulation}


\subsection{Preliminaries}
\textbf{Markov Decision Process (MDP):} An MDP is a fundamental framework defined by a tuple comprising four essential elements: $(s_{t}, a_{t}, p(s_{t+1}|s_{t}, a_{t}), r(s_{t+1}|s_{t}, a_{t}))$. Here, $s_t$ and $a_t$ represent the continuous state and action at time step~$t$, respectively. The function $p(s_{t+1}|s_{t}, a_{t})$ signifies the transition probability, indicating the likelihood of transitioning from the current state $s_{t}$ to the next state $s_{t+1}$ given the action $a_t$. The function $r(s_{t+1}|s_{t}, a_{t})$ denotes the immediate reward received from the environment.

\textbf{Off-policy RL:} In online RL, an agent engages in continuous interactions with the environment to gather experiences and learn the optimal policy $\pi^*$. The objective is to maximize the expected future return $R_{t}=\mathbb{E}[\sum_{i=t}^\infty\gamma^{i-t}r_{i+1}]$, with a discount factor $\gamma \in [0, 1]$ accounting for the importance of future rewards. The expected return under a policy $\pi$ after taking action $a$ in the state $s$ is computed by the corresponding action-value function $Q^{\pi}(s, a)=\mathbb{E}[R_t|s_t=s, a_t=a]$. This can be computed using the Bellman equation:
\begin{equation}\label{Qfunc}
\begin{split}
\scalemath{0.87}{
Q^{\pi}(s_t, a_t)  = \mathbb{E}_{s_{t+1}\sim p}[r(s_t, a_t)+ \gamma\mathbb{E}_{a_{t+1}\sim{A}}[Q_{\pi}(s_{t+1}, a_{t+1})]],
}
\end{split}
\end{equation}

\noindent where $A$ denotes the action space. The ultimate goal of RL algorithms is to discover an optimal policy $\pi^*$ such that $Q^{\pi^*}(s, a)=Q^{*}(s, a)$ for all states and actions. 

\subsection{Problem Formulation}
Initially, the target object remains ungraspable due to surrounding objects. Given the initial pose of the target object, $p_s$, the push policy aims to determine the proper parameters (i.e., initial push point, direction, and length) to manipulate the scene, making the target object graspable (i.e., the grasp quality of the target object exceeds a predefined threshold). In essence, the push policy is strategically conditioned upon the output of an object-agnostic grasping policy, 
\begin{wrapfigure}{r}{0.5\linewidth}
    \includegraphics[width=\linewidth, trim={1cm, 0.5cm, 1.2cm, 0cm}, clip=true]{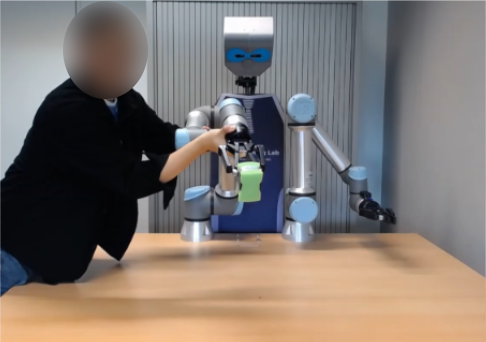}
    \caption{Kinesetic teaching of throwing kernel.}
    \label{fig:kinesthetic_teaching}
    \vspace{-3mm}
\end{wrapfigure}
which has undergone unsupervised training. Similarly, the throwing policy seeks to determine the appropriate residual parameters of the initial throwing kernel, previously taught to the robot through Kinesthetic teaching, while considering the goal pose $p_g$ (i.e., the pose of the basket) in the environment (see Fig.~\ref{fig:kinesthetic_teaching}).
These tasks can be represented using a fully observable MDP framework, and an off-policy reinforcement learning framework is employed to solve them. It is important to highlight that each of these policies is trained independently. This decision is grounded in the understanding that achieving end-to-end training for such multifaceted tasks would necessitate extensive training samples. Notably, the push policy is strategically reliant on the output of an object-agnostic grasping policy. This modular approach enhances the system's decomposability, thereby enabling efficient troubleshooting, fine-tuning, and updates for each policy, all without causing ripple effects throughout the entire system.

\subsubsection{\textbf{States}}

In both the push and throw policies, we utilize a feature vector to characterize the continuous state. In the context of the push and grasp policy, the observation space includes a flattened grasp quality map consisting of $50\times50$ pixels. Additionally, it encompasses eleven supplementary parameters, comprising the initial push point coordinates \mbox{(X, Y, Z)}, push direction, and push length, all represented by five floating-point numbers. Further, it includes the coordinates of the target object (X, Y, Z), and three float numbers representing the initial grasp quality of the target object, the grasp quality of the target object after a push, and the difference between these qualities.

Within the throwing policy, the observation state encompasses several critical elements. These include the robot's proprioception, information regarding the target goal, releasing time, duration of trajectory execution, and the distance between the thrown object and the goal. 
The robot autonomously learns the initial and final values for the shoulder joint, the duration of trajectory execution (speed), and the optimal releasing time during the learning process.

At each time step $t$, we record essential parameters. This includes the initial and final shoulder joint values ($j_i$ and $j_f$) in radians, representing proprioception as \mbox{${\rm proprio}=(j_i, j_f) \in \mathbb{R}^2$}. The position of the goal (center of the box) in the task space is described as ${\rm goal}=(x^g, y^g, z^g) \in \mathbb{R}^3$. Additionally, we consider the absolute distance of the thrown object relative to the goal, including distances in the X and Y axes. These distances are represented as ${\rm dist}=(d^g, d^g_x, d^g_y) \in \mathbb{R}^3$. Furthermore, we record two timing profiles: the duration of executing the throwing trajectory $\tau$, and the time for releasing the object, $t_r$, where $t_r < \tau$. These timings are captured as ${\rm time} = (t_r , \tau) \in \mathbb{R}^2$. In summary, the state space of throwing policy can be represented as a vector: \mbox{$s=({\rm proprio}, {\rm goal}, {\rm dist}, {\rm time}) \in \mathbb{R}^{10}$}.

\subsubsection{\textbf{Actions}}

In the case of the push and grasp policy, the action space is defined as follows: A grasp action is taken when the maximum grasp quality of the target object exceeds a threshold, where the threshold is set at 0.7, otherwise, a push action is executed. For the throwing policy, each action is denoted by a vector \mbox{$a \in ([-1, 1])^4$}, which represents (\textit{i})~the initial and (\textit{ii})~the final shoulder joint values, (\textit{iii})~the duration of trajectory execution, and (\textit{iv}) the releasing time.

\subsubsection{\textbf{Transition function}}
During each training episode, we introduce random variations to the positions of the basket, the target object, and the surrounding objects in a manner that renders the target object initially ungraspable. Consequently, the dynamics of the transition function are influenced by the execution of push actions or throwing trajectories, based on the current state of the scene and sampled parameters. To be specific, we compute the subsequent state, denoted as $s_{i+1}$, following the execution of the action $a_i$, as defined by the transition function $f_s$; in mathematical terms, this relationship is expressed as $s_{i+1} = f_s(s_i, a_i)$, where $s_{i}$ represents the current state, and $a_{i}$ signifies the action undertaken. It should be noted that, given the unknown nature of the transition function, our off-policy reinforcement learning framework remains model-free, adapting to various scenarios without requiring a predefined model.

\subsubsection{\textbf{Rewards}}

Our objective for the push and grasp policy is to equip the robot with the capability to efficiently manipulate the scene so that the target object becomes graspable following a push action. This is indicated by the grasp quality of the target object surpassing a predefined threshold, denoted as $\tau_g=0.7$, for successful grasping. For successful grasping, we assign a maximum reward of $1.0$. Additionally, if the grasp quality of the target object improves after the execution of a push action, we compute the rewards as follows: $R = \alpha \cdot e^{(-d^2/0.001)} + (1-\alpha) \cdot e^{(-d^2/0.05)}$. Here, $\alpha$ is set to $0.9$, and $d$ is calculated as $1 - \beta$, with $\beta$ representing the maximum grasp quality of the target object. On the other hand, actions deemed as ineffective or useless are penalized with a reward of $-0.1$. Terminal conditions for the push and grasp policy include:
\begin{itemize}
    \item The target object goes beyond the robot's workspace.
    \item The number of consecutive push actions exceeds a predefined budget (i.e., set at five consecutive pushes).
    \item The robot executes a grasp action.
\end{itemize}

In the case of the throwing action, a successful outcome is achieved if the thrown object lands inside the target basket ($R=1.0$).
We determine success by calculating the absolute distance between the object and the goal, where the distance is compared to the radius (${\rm d}$) of a cylindrical space fitted inside the basket. If the next state results in success (i.e., the thrown object lands inside the basket), we incentivize this behavior by setting the reward $R=1.0$. Conversely, if the throwing action results in the object landing outside the target basket, the reward is penalized based on the distance, with $R=-{\rm dis}({\rm o}, g)$. A throwing episode is terminated upon the execution of a throwing action.

\subsection{Perception}
\begin{figure}[!t]
\centering
\vspace{-5mm}
\includegraphics[width=\linewidth,trim={0cm, 0.8cm, 0cm, 0cm}, clip=true]{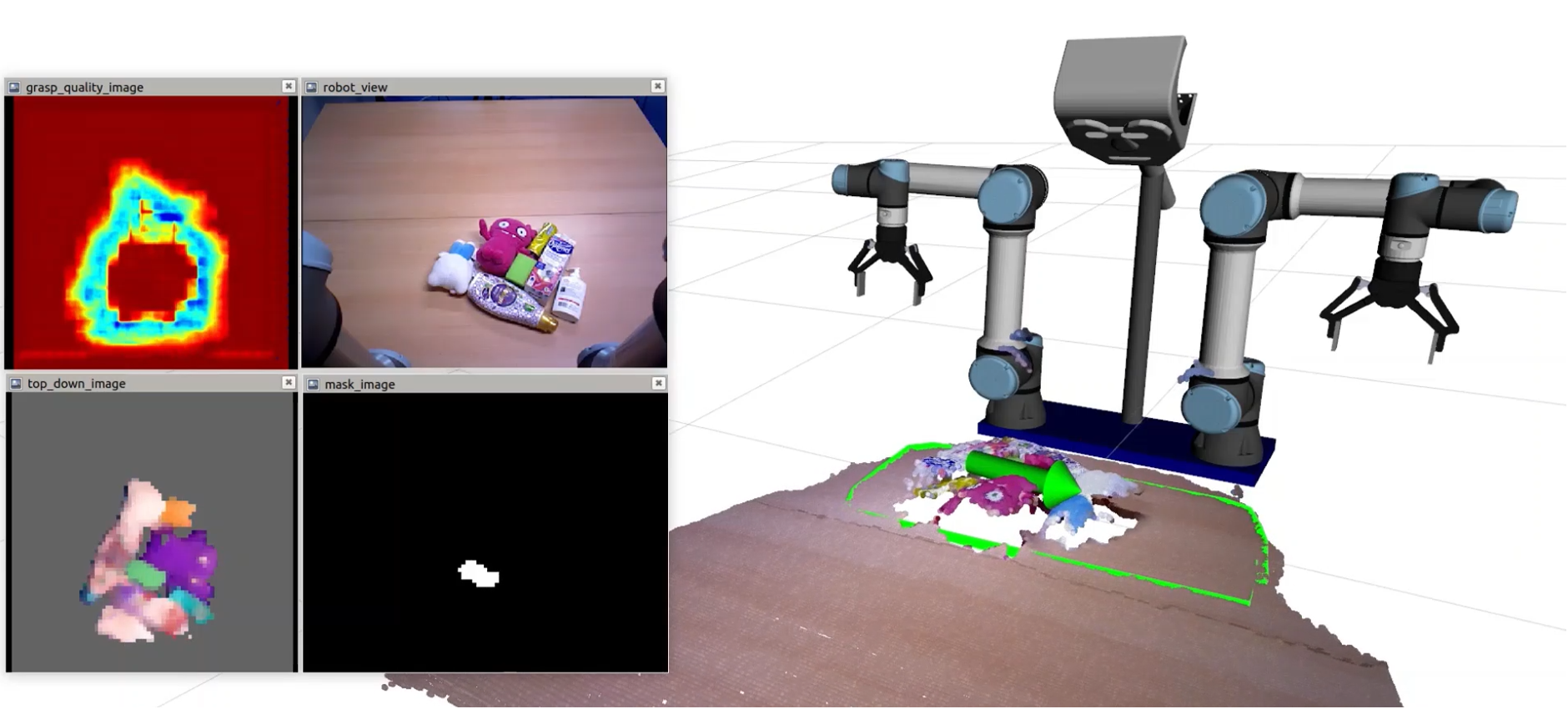}
\caption{Visualizing the output of our perception system: world model information is provided through a top-down view, a grasp quality map, and a mask of the target object. The robot's workspace is outlined by the green rectangle, and the predicted push action is shown by the green arrow.
}
\label{fig:perception}
\end{figure}
\begin{figure*}[!t]
\centering
\includegraphics[width=\linewidth, trim={0cm, 0cm, 0cm, 0cm}, clip=true]{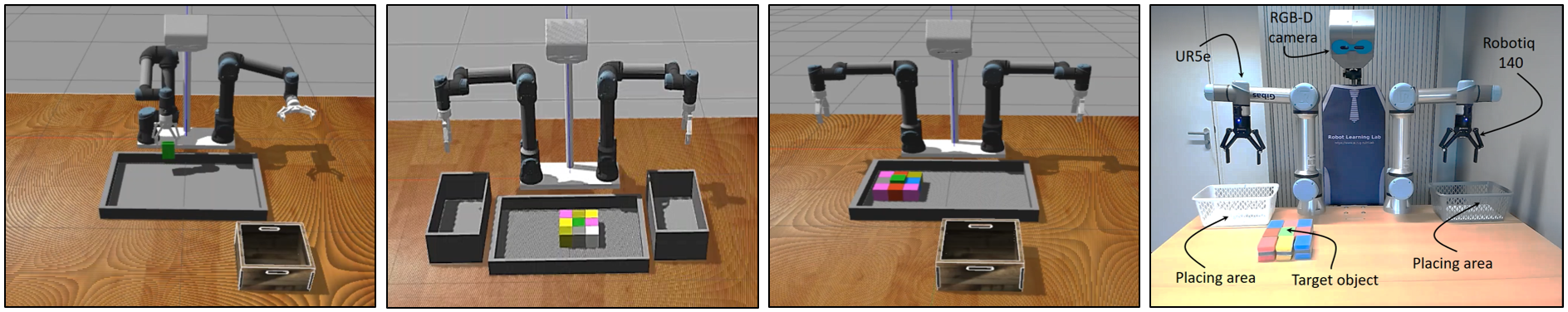}
\caption{Our experimental setups: (\textit{from left to right}) Training the throwing policy, the push-and-grasp policy, integrating all policies into a unified robotic system, and the real dual-arm robot setup.}
\vspace{-4mm}
\label{fig:simulation_and_real_robot}
\end{figure*}
In the interest of safety, our push, grasp, and throw policies were initially trained in a simulated environment before being tested on our real-robot platform. To facilitate this transition, we developed a versatile interface capable of processing \mbox{RGB-D} sensory data, which was utilized seamlessly in both our simulation and real robot experiments \cite{kasaei2019interactive}\cite{kasaei2021simultaneous}\cite{kasaei2023mvgrasp}. Specifically, we employed an \mbox{RGB-D} Asus Xtion camera, capturing point cloud data at a rate of 30 Hz. 
For tracking the position of the target object, we adopted a straightforward object detection method that takes into account both shape and color data~\cite{kasaei2018towards}. It is important to note that, we constrained the system to grasp the object from above and near its center of mass~\cite{kasaei2023mvgrasp}. Our perception system serves as a world model service, enabling the agent to access real-time environmental information at each time step. Figure~\ref{fig:perception} provides a visual representation of the outcomes generated by our perception system. For a comprehensive understanding of our perception and grasping pipelines, we encourage referring to our earlier research for more detailed insights~\cite{kasaei2018towards}\cite{kasaei2023mvgrasp}. 





\section{Experiments and Results}
Our method underwent extensive experimentation in both simulated and real-robot scenarios to validate its efficacy. This evaluation involves assessing the proposed approach based on several key performance metrics: (i) \textit{Success Rate} calculating by dividing the number of instances in which the robot successfully places the target object into the basket by the total number of trials. (ii) \textit{Required Number of Actions} needed to successfully singulate the target object.


\subsection{Experimental setup and tasks settings}

Our experimental setups in both simulation and real-robot environments are represented in Fig.~\ref{fig:simulation_and_real_robot}. Specifically, we developed a simulation environment in Gazebo, employing the ODE physics engine to closely emulate our real robot. Our setup includes an Asus Xtion camera, two Universal Robots (UR5e) equipped with Robotiq 2F-140 grippers, and an interface for initiating and concluding experiments.
To evaluate the proposed approach, we devised three distinct tasks, each progressively more challenging:
\begin{itemize}
\item \textbf{Task 1:} The objective here is to singulate the target object from a cluttered scene and place it into a reachable basket. An illustrative example of this task in a real-robot setup can be seen in Fig.~\ref{fig:simulation_and_real_robot} (\textit{center}).

\item \textbf{Task 2:} The robot is tasked with throwing an object into a basket positioned randomly in front of it. Fig.~\ref{fig:simulation_and_real_robot} (\textit{left}) offers a glimpse of this task in a simulation environment.

\item \textbf{Task 3:} This multifaceted task requires the robot to first singulate the target object, followed by grasping and throwing it into a basket situated beyond the robot's maximum kinematic range. A visual depiction of this task is shown in Fig.~\ref{fig:simulation_and_real_robot}  (\textit{right}).
\end{itemize}

For each of the proposed tasks, we trained the model for $100,000$ steps in the simulation. Specifically, we trained the throwing policy using a cubic object as shown in Fig.~\ref{fig:simulation_and_real_robot} (\textit{left}). Moreover, we conducted the training of the push and grasp policy in a cubic scenario, where the target object initially resides in a configuration that renders it ungraspable, as illustrated in Fig.~\ref{fig:simulation_and_real_robot} (\textit{center}). Subsequently, we evaluated the learned policy on real robots across ten distinct scenarios, as depicted in Fig.~\ref{fig:real_cubic_scenarios}. 

\subsection{Baseline Methods}
\begin{wraptable}{r}{0.55\linewidth}
    \vspace{-4mm}
    \centering    
    \caption{SAC hyper-parameters}
    \resizebox{0.95\linewidth}{!}{
    \begin{tabular}{|c|c|}
    \hline
    \textbf{Parameter} & \textbf{Value} \\
    \hline
\#hidden layers (all networks) &  $2$ \\
\#hidden units per layer & $256$\\
\#samples per minibatch & $256$\\
optimizer & Adam\\
learning rate  & $3 \times 10^{-4}$\\
batch size & 256 \\
\#epochs & 50K \\
discount ($\gamma$)  & $0.99$\\
replay buffer size &  $10^6$\\
nonlinearity & ReLU\\
target update rate ($\tau$) & 0.005\\
target update interval & 1 \\
gradient steps & 1 \\\hline
    \end{tabular}}
    \label{tab:SAC_hyper}
    \vspace{-3mm}
\end{wraptable}
We harnessed two state-of-the-art, sample-efficient off-policy RL algorithms to train our robotic system: Deep Deterministic Policy Gradient~(DDPG)~\cite{lillicrap2015continuous,silver2014deterministic}, and Soft Actor-Critic~(SAC)~\cite{haarnoja2018soft,haarnoja2018soft2} via the stable baseline ~\cite{stable-baselines3}.
The neural network architectures for both SAC and DDPG consisted of two hidden layers, each comprising 256 neurons, activated by Rectified Linear Units (ReLU) activation functions. The hyper-parameters of SAC are listed in Table.~\ref{tab:SAC_hyper}, and the hyper-parameters of DDPG that are not listed in Table~\ref{tab:SAC_hyper} are reported in the Experiments section of our previous work~\cite{luo2020accelerating}.

\subsection{Results}

\begin{figure}[!t]
    \centering
    \includegraphics[width=\linewidth]{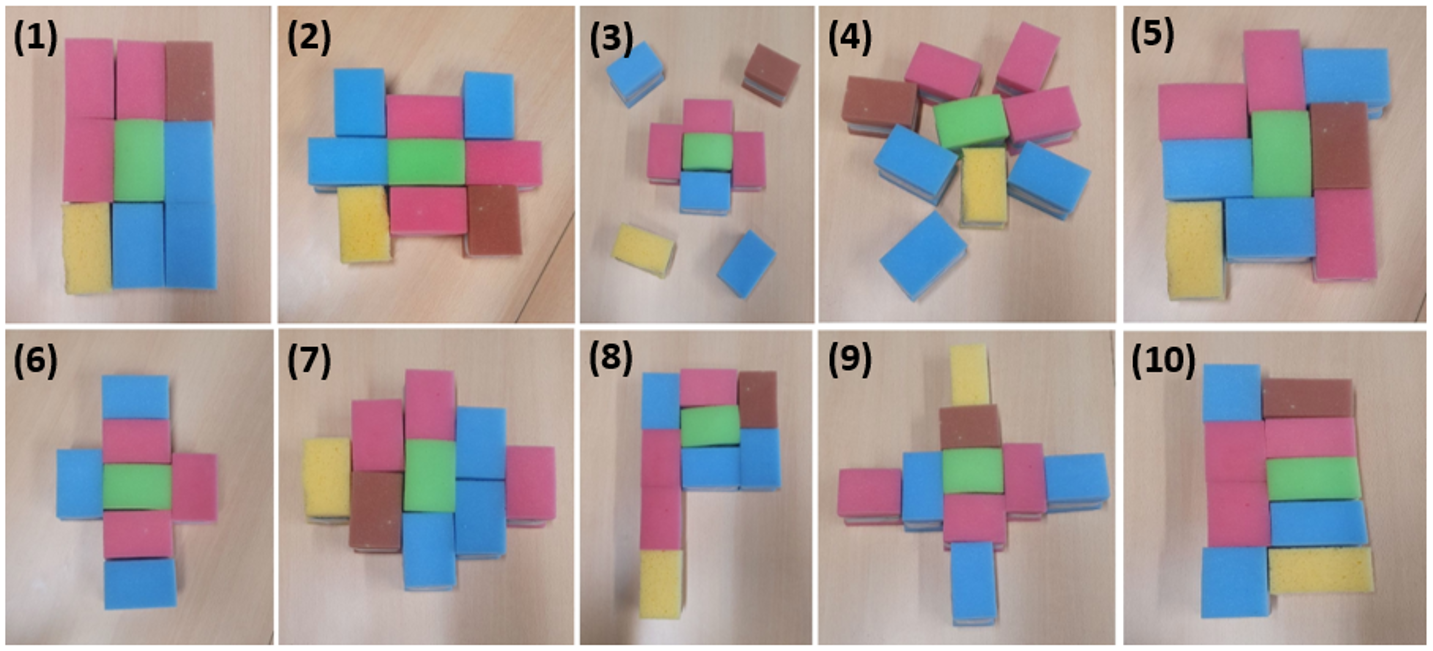}
    \caption{We created 10 cubic scenarios to assess the efficacy of the acquired policies in real robot experiments.}
    \label{fig:real_cubic_scenarios}  
    \vspace{-4mm}
\end{figure}

In the context of \textbf{Task 1}, which involves singulating the target object from a cluttered scene and depositing it into a reachable basket, we conducted a comprehensive evaluation comprising 1000 experiments with a cubic scenario, 500 simulation experiments with the SAC policy, 500 simulation experiments with the DDPG policy. Furthermore, we executed 200 real-world experiments, spanning across 10 distinct cubic scenarios (see Fig.~\ref{fig:real_cubic_scenarios}). Notably, the first scenario closely resembled the simulated environment (\#1), while the remaining nine scenarios presented entirely novel challenges that had not been encountered during training. For each scenario, we executed 10 real robot tests with the SAC policy and 10 real robot experiments with the DDPG policy. This round of experiments provides key insights into the comparative efficacy and efficiency of the two algorithms. Table~\ref{tab:results_task1} summarizes the outcomes.

\begin{table}[!b]
    \centering
    \vspace{-4mm}
    \caption{Performcane of the SAC and DDPG policies on 10 real scenarios in Task 1.}
     \resizebox{\linewidth}{!}{
    \begin{tabular}{|c|c|c|c|c|}
      \hline
       \multirow{2}{*}{\textbf{Scenario}} &\multicolumn{2}{c|}{\textbf{Task Success Rate (↑)
}}&\multicolumn{2}{c|}{\textbf{Avg. Number of Actions (↓)}}\\\cline{2-5} 
       & \textbf{SAC} & \textbf{DDPG} & \textbf{SAC} & \textbf{DDPG} \\\hline
       \#1  & 100    &  80 & 1.6 & 2.7  \\\hline
       \#2  & 100    &  90 & 1.8 & 3.4  \\\hline
       \#3  & 90     &  90 & 2.4 & 4.3  \\\hline
       \#4  & 90     &  100 & 2.9 & 3.8  \\\hline
       \#5  & 100    &  90 & 2.8 & 3.3  \\\hline
       \#6  & 100    &  100 & 1.7 & 2.8  \\\hline
       \#7  & 100    &  80 & 3.2 & 4.2  \\\hline
       \#8  & 90     &  90 & 3.3 & 4.4  \\\hline
       \#9  & 80     &  70 & 3.8 & 4.8  \\\hline
       \#10  & 90    &  70 & 3.1 & 3.8  \\\hline
       avg $\pm$ std & \textbf{94 $\pm$ 6.63}  &  86 $\pm$ 10.19 & \textbf{2.66 $\pm$ 0.71} & 3.75 $\pm$ 0.66  \\\hline
    \end{tabular}}
    
    \label{tab:results_task1}
\end{table}

\begin{figure}[!t]
    \centering
\includegraphics[width=\linewidth]{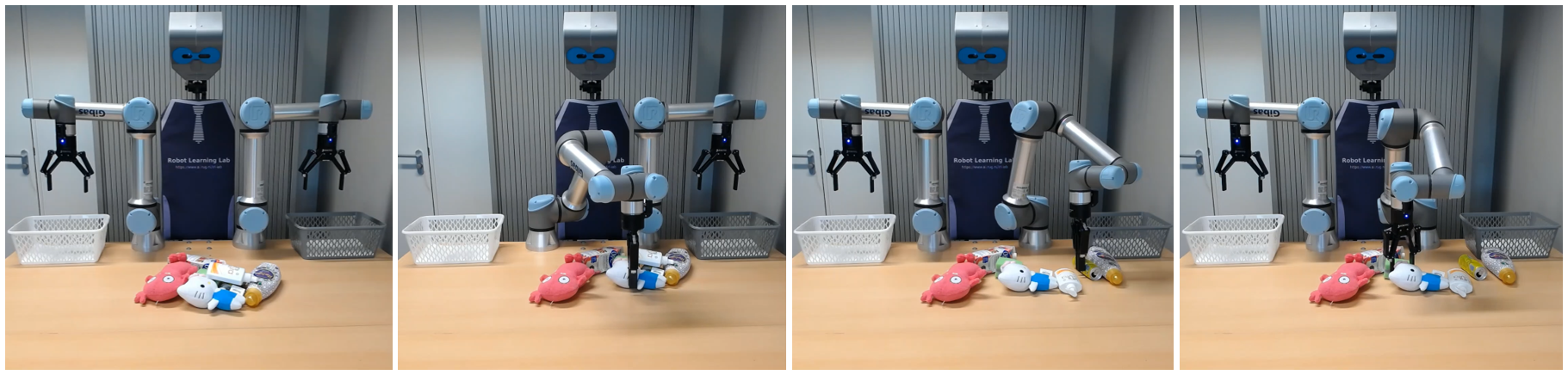}
\caption{A real robot experiment demonstrating the singulation and grasping of a fully visible target object (a green sponge) from a pile of household objects. 
}
\label{fig:pile_of_object}
\vspace{-5mm}
\end{figure}

By comparing the obtained results, it is clear that under the SAC policy, the task success rate (percentage of successful trials) ranged from 80\% to 100\%, while under the DDPG policy, it varied from 70\% to 100\%. The average number of actions required to accomplish the task (lower is better) ranged from 1.6 to 3.8 for SAC and 2.7 to 4.8 for DDPG. Overall, the results indicate that the SAC policy outperformed DDPG in most scenarios, achieving a higher task success rate and requiring fewer actions on average to complete the task. On average, SAC achieved a task success rate of 94\% with a standard deviation of approximately 6.63\%, while DDPG achieved an average success rate of 86\% with a standard deviation of about 10.19\%. Additionally, SAC exhibited an average number of actions of approximately 2.66 with a standard deviation of 0.71, whereas DDPG had an average of about 3.75 actions with a standard deviation of approximately 0.66. These results suggest that the SAC policy is more robust and effective in handling the complexity of real-world scenarios, as it consistently outperformed DDPG across various scenarios in \textbf{Task 1}. 
\begin{figure}[!b]
\centering
\vspace{-4mm}
\includegraphics[width=\linewidth]{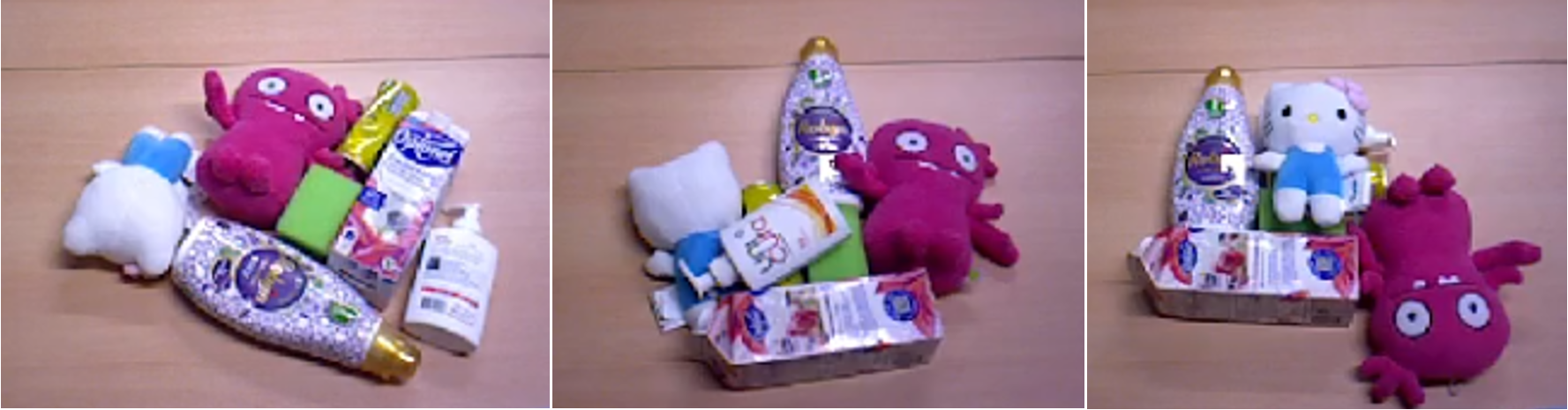}
\caption{
Visibility of the target object (green sponge) in Task 1 : (\textit{left}) fully visible, (\textit{center}) partially visible, (\textit{right}) completely concealed.
}
\vspace{-4mm}
\label{fig:task1_pile_scenarios}
\end{figure}
Furthermore, we evaluated the learned policies using various piles of objects scenarios. We used $10$ daily-life objects with different materials, shapes, sizes, and weights. In these scenarios, the initial visibility of the target object was manipulated to create three distinct conditions: (i) the target object was fully visible, (ii) it was partially visible, and (iii) it was entirely concealed. To provide clarity, ~Fig.~\ref{fig:task1_pile_scenarios} showcases a representative example of each of these scenarios. 
Similar to the previous round of experiments, for each scenario we performed 20 real robot experiments including 10 experiments with SAC and 10 experiments with DDPG. Results are reported in Table~\ref{tab:results_task1_pile}. Fig.~\ref{fig:pile_of_object} shows an example of a fully visible scenario.

By comparing all results, it is clear that in the ``\textit{fully visible}'' scenario, where the target object was completely visible initially, the SAC policy achieved a 100\% task success rate, while the DDPG policy achieved an 80\% success rate. The SAC policy required an average of 2.9 actions, whereas DDPG needed an average of 3.3 actions to complete the task. In the ``\textit{partially visible}'' scenario, SAC achieved a 100\% task success rate, while DDPG achieved an 80\% success rate. SAC had an average action count of 3.1, while DDPG averaged 3.6 actions. In the ``\textit{not visible}'' scenario, where the target object was entirely hidden from view initially, the SAC policy achieved an 80\% task success rate, while DDPG reached 60\%. SAC required an average of 3.5 actions, while DDPG needed an average of 4.7 actions to accomplish the task. These results demonstrate the robustness of the SAC policy across different initial visibility conditions. In scenarios where the target object was partially or fully visible, both policies achieved high success rates. However, the SAC policy consistently required fewer actions to complete the task. In the challenging ``\textit{not visible}'' scenario, SAC outperformed DDPG, highlighting its effectiveness in handling complex real-world situations.

\begin{table}[!b]
    \centering
    \vspace{-4mm}
    \caption{Performcane of the robot on piles of real objects.}
     \resizebox{\linewidth}{!}{
    \begin{tabular}{|c|c|c|c|c|}
      \hline
       \multirow{2}{*}{\textbf{Scenario}} &\multicolumn{2}{c|}{\textbf{Task Success Rate (↑)
}}&\multicolumn{2}{c|}{\textbf{Avg. Number of Actions (↓)}}\\\cline{2-5} 
       & \textbf{SAC} & \textbf{DDPG} & \textbf{SAC} & \textbf{DDPG} \\\hline
       fully visible  & 100    &  80 & 2.9 & 3.3  \\\hline
       partially visible  & 100    &  80 & 3.1 & 3.6  \\\hline
       not visible  & 80     &  60 & 3.5 & 4.7  \\\hline
    \end{tabular}}
    
    \label{tab:results_task1_pile}
\end{table}

In the second round of experiments (\textbf{Task2}), which involves the task of throwing a target object into a basket, the SAC policy demonstrated superior performance when compared to the DDPG policy. In the simulated environment, the robot achieved a success rate of $95\%$ with the SAC policy and $92\%$ with the DDPG policy. However, in the real-robot experiments, the differences in performance became more pronounced. The robot equipped with the SAC policy achieved remarkable results, boasting a success rate of $92\%$. In contrast, when utilizing the DDPG policy, the robot's throwing performance experienced a decline, with a success rate of $81\%$. This observed disparity between SAC and DDPG in real-robot experiments, as opposed to the relatively similar performance observed in simulation, underscores the adaptability and robustness of the SAC in addressing the complexities inherent in real-world scenarios.

\textbf{Task 3} represents a significantly more complex challenge compared to the previous tasks, requiring the robot to sequentially singulate the target object, grasp it, and then execute a throw into a basket positioned beyond the robot's maximum kinematic range. As anticipated, the robot equipped with the SAC policy consistently outperformed its counterpart using the DDPG policy in both simulation and real-robot experiments. Specifically, the robot with the SAC policy achieved an impressive $88\%$ success rate while requiring an average of $1.8$ push actions per episode. Conversely, the robot using the DDPG policy achieved a success rate of $79\%$ but required an average of $2.7$ push actions per episode. Transitioning to the real-robot experiments, the performance gap between the two policies persisted, albeit with some modifications. The robot equipped with the SAC policy maintained a relatively high success rate of $85\%$ while slightly increasing the average number of push actions to $2.1$. Similarly, the robot utilizing the DDPG policy experienced a drop in success rate to $70\%$ and a more significant increase in the average number of push actions to $4.4$. 

Comparing the performance across all tasks, it becomes evident that the difference in performance between the SAC policy and alternative policies becomes more pronounced as the task complexity increases. This disparity is particularly evident in the real-robot experiments. 

\subsection{Failure Cases}

Throughout our experimental trials, we encountered four distinct categories of failures, they are including:

\noindent\textbf{Out of workspace failure:} In some real-robot experiments, the robot unintentionally moved the target object out of the workspace. This type of failure underscores the importance of maintaining objects within the robot's operational boundaries. Addressing such occurrences can lead to improved control strategies to prevent objects from being pushed out of the workspace. It is worth noting that such failures were infrequent in the simulation environment, primarily because we imposed spatial constraints by encapsulating objects within a box-like workspace. 

\noindent\textbf{Inaccurate grasp pose failure:} Another failure scenario arose when the predicted grasp pose was inaccurate. This led to instances where the robot either failed to grasp the target object or mistakenly grasped multiple objects simultaneously. Addressing this challenge calls for enhanced perception systems and grasp pose estimation techniques. 

\noindent\textbf{Pushing limitation failure:} There were cases where the robot could not make the target object graspable even after the maximum allowable number of pushes. This indicates that the policy might need refinement, particularly in scenarios with complex object arrangements. Developing strategies to effectively manipulate cluttered scenes, such as learning more sophisticated non-prehensile pushing behaviors, can be essential for overcoming this limitation.

\noindent\textbf{Throwing failure:} The primary factors contributing to the failures observed in the throwing policy were the imprecise prediction of parameters and the latency in executing the gripper commands. Specifically, the gripper's control process entails an intermediate step, where commands are relayed through the robot's controller before reaching the gripper itself. This intermediary layer introduces a potential delay, influenced by network status and the robot's conditions.


\vspace{-1mm}
\section{CONCLUSIONS}

In this work, we addressed the intricate challenges of harnessing the synergy between pushing, grasping, and throwing actions to enhance object manipulation in cluttered scenarios. 
We decoupled the learning of pushing, grasping, and throwing policies, recognizing that end-to-end training for such multifaceted tasks demands an extensive corpus of training samples. 
The push policy strategically relies on the output of an object-agnostic grasping policy.  This modular approach enhances the system's decomposability, enabling streamlined troubleshooting, fine-tuning, and updates to each policy without cascading effects on the entire robotic system. 
Through extensive experiments in both simulation and real-robot settings, we evaluated the performance of our approach across tasks of ascending complexity. Our results demonstrated that the SAC policies learned in simulation effectively transferred to the real-robot setup, showcasing impressive generalization capabilities to new target locations and unseen objects. We achieved success rates of over 80\% in both simulation and real-robot environments, underlining the effectiveness of our approach. 
However, we also encountered challenges and observed failures. These included unintentional movements of the target object out of the robot's workspace, inaccuracies in grasp pose predictions leading to failed grasps or grasping multiple objects, and difficulties in making the target object graspable with a limited number of pushes. 
In future work, we will leverage these insights, in combination with the integration of common-sense knowledge through Large Language Models (LLMs), to handle sophisticated household tasks. 

\section{Acknowledgements} 
We thank the Center for Information Technology of the University of Groningen for their support and for providing access to the Hábrók high-performance computing cluster. This work is partially supported by Google DeepMind through the Research Scholar Program for the ``\textit{Continual Robot Learning in Human-Centered Environments}'' project. 


{
\small
\bibliographystyle{IEEEtran}
\bibliography{refs}
}

\end{document}